# Causal Rule Forest: Toward Interpretable and Precise Treatment Effect Estimation

Chan Hsu
Department of Information Management
National Sun Yat-sen University
Kaohsiung, Taiwan
chanshsu@gmail.com

Jun-Ting Wu
Department of Information Management
National Sun Yat-sen University
Kaohsiung, Taiwan
lydiazwu@gmail.com

Yihuang Kang
Department of Information Management
National Sun Yat-sen University
Kaohsiung, Taiwan
ykang@mis.nsysu.edu.tw

*Abstract*—Understanding and inferencing Heterogeneous Treatment Effects (HTE) and Conditional Average Treatment Effects (CATE) are vital for developing personalized treatment recommendations. Many state-of-the-art approaches achieve inspiring performance in estimating HTE on benchmark datasets or simulation studies. However, the indirect predicting manner and complex model architecture reduce the interpretability of these approaches. To mitigate the gap between predictive performance and heterogeneity interpretability, we introduce the Causal Rule Forest (CRF), a novel approach to learning hidden patterns from data and transforming the patterns into interpretable multi-level Boolean rules. By training the other interpretable causal inference models with data representation learned by CRF, we can reduce the predictive errors of these models in estimating HTE and CATE, while keeping their interpretability for identifying subgroups that a treatment is more effective. Our experiments underscore the potential of CRF to advance personalized interventions and policies, paving the way for future research to enhance its scalability and application across complex causal inference challenges.

## I. Introduction

Evaluating the effectiveness of interventions is vital in various domains such as healthcare and economics. Typically, new medications are subjected to Randomized Control Trials (RCTs) to validate their safety and efficacy within the targeted population. However, RCTs can be impractical or unethical due to high costs or potential harm, necessitating the use of observational data to retrospectively estimate causal effects. To mitigate biases introduced by non-random treatment assignments, statistical methods like Propensity Score Matching (PSM) are utilized to align samples according to their probability of receiving treatment [1]. As the demand for personalized decision-making increases, identifying heterogeneity in treatment effects across different subgroups and individuals becomes critical. Machine Learning (ML) approaches are employed to uncover complex interactions among covariates, enabling more precise estimation [2], [3].

These causal ML models, ranging from decision trees to Deep Neural Networks (DNN) [4], [5], [6], [7], address a principal challenge in estimating Heterogeneous Treatment Effects (HTE): the absence of ground truth. Since we cannot observe the outcomes with and without treatments simultaneously in the same individual, most causal ML approaches rely on Rubin's potential outcome framework to estimate treatment effects. Another significant issue in estimating HTE with observational data is biased estimation introduced by confounder variables and their intricate interaction that causes the non-random treatment assignment in samples with varying characteristics.

Benefiting from the advances in computing power and deep representation learning [8], causal ML models are capable of learning complex representations of interaction and confounding for all covariates [3]. The approaches either developed based on DNN or following the indirect estimation manner, have demonstrated remarkable performance in simulation studies and benchmark datasets [6], [7], [9], [10]. However, their complex architecture and indirect estimations reduce the interpretability of treatment effect heterogeneity, raising concerns about the reliability of these models despite their superior performance over more interpretable alternatives.

To bridge the gap between the capability of learning more complex patterns and the demand for interpretable insights into treatment effect heterogeneity, we introduce the Causal Rule Forest (CRF). Inspired by the concept of Deep Rule Forest (DRF) [11], the CRF augments causal trees with rule-encoded data representations to refine the HTE and Conditional Average Treatment Effect (CATE) estimation. Employing multi-level Boolean rules in Disjunctive Normal Form (DNF), such as "IF (smoking AND hypertension) OR (obesity AND diabetes) THEN CATE = 2", the CRF enhances the predictive performance of Causal Trees while preserving their interpretability. This approach enables more nuanced subgroup interaction representations, facilitating personalized decision-making and deepening our understanding of complex treatment effects. The contributions of CRF are multifaceted:

- It provides a flexible, deep model architecture that elevates the performance of existing interpretable models through advanced rule-based data representations.

- It optimizes local CATE estimations, allowing for more precise descriptions of the causal relationships between treatments and outcomes.

- It fosters the learning of interpretable causal data representations using multi-level IF-THEN Boolean logic, aiding in elucidating causality and the potential correction of algorithmic biases.

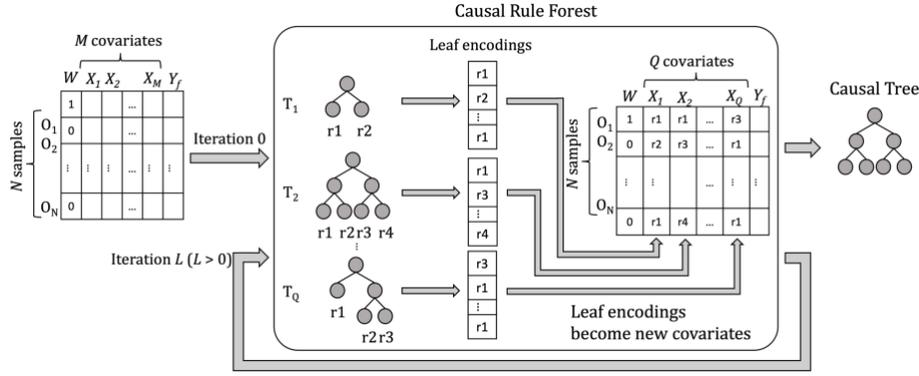

Fig. 1. Framework of Causal Rule Forest

The remainder of this paper is structured as follows: Section 2 introduces causal inference models to HTE/CATE estimations. Section 3 defines the problem and details the proposed CRF. Section 4 presents and discusses the experimental results, including performance comparison in HTE/CATE estimations and the utilization of rule representations for robust subgroup discovery. Section 5 concludes our work

## II. BACKGROUND

Estimating the treatment effects of interventions is crucial for evaluating the effectiveness of treatments or policies in various domains such as healthcare and economics. Unlike typical supervised learning tasks, where ground truth for classification and regression is available, treatment effect estimation is challenged by the inability to observe outcomes under both treatment and control conditions simultaneously in the real world. Therefore, most causal ML models adopt Rubin's potential outcome framework, which integrates observed and hypothesized counterfactual outcomes to estimate treatment effects [12].

In the realm of machine learning models for estimating HTE and CATE, methodologies are generally divided into direct and indirect estimators. Direct estimators utilize input features to determine the treatment effect for individual samples. In contrast, indirect estimators first predict potential outcomes with and without the treatment and subsequently compute the treatment effect as the difference between these two outcomes [13].

As a majority in earlier literature, direct estimators like Causal Inference Trees [4] and Causal Trees [5] were proposed to identify confounder variables managing their impact on treatment effect estimation. Causal Inference Trees enhance decision-making by incorporating facilitating scores to mitigate the impact of confounders through the partitioning process, aiming for a more uniform treatment effect estimation across nodes. Causal Trees adopt an approach of 'honest estimation', using separate data subsets for growing the trees and estimating CATE, thereby minimizing the risk of bias in treatment effect estimation. More recently, Casual Interaction Trees [14] have been introduced, which refine the control of confounders using methods like the inverse probability of treatment weighting and the g-formula, offering a robust estimation of HTE that improves upon earlier models such as Interaction Trees.

The introduction of ensemble methods like Causal Forest [15] and Generalized Random Forest [16] marks a significant advance, reducing the variance in HTE estimations by aggregating multiple trees' outputs and the error in treatment effect estimation. Although these methods enhance precision, they typically do so at the expense of interpretability, particularly in understanding treatment effect heterogeneity. Concurrently, the exponential growth in computational capabilities has facilitated the development of causal ML models based on deep learning, which excel at discerning complex patterns and interactions obscured by confounding variables [10]. This category of models, primarily comprising indirect estimators, is increasingly favored for its superior performance, though it often suffers from reduced interpretability compared to more straightforward models [6], [17], [18].

Despite the progress in modeling techniques, a critical issue persists the trade-off between model performance and interpretability. The lack of clarity on how ensemble and indirect estimators derive their estimations raises concerns about their reliability, particularly in high-stakes decision-making contexts [19]. In response, there has been a growing interest in developing interpretable ML approaches. Recent innovations focus on learning decision rule sets that users can readily understand, enhancing transparency [20], [21], [22]. Our study builds on these efforts by implementing multi-level Boolean logic in rule learning, enhancing both the interpretative depth and predictive performance of these rule sets, as exemplified by our proposed CRF model

## III. TREATMENT EFFECT ESTIMATION USING CAUSAL RULE FOREST

In this section, we introduce the Causal Rule Forest (CRF) for estimating HTE and identifying subgroup heterogeneity in treatment effects. The CRF leverages the concept of stacking multiple layers of models, similar to techniques used in deep learning, enabling it to discover more complex causal representations from data. We begin by outlining the assumptions of the CRF and defining the challenges in estimating treatment effects, followed by a detailed description of how a CRF model is constructed.

**Algorithm 1: Building Causal Rule Forest**

1. **Input**: $Z_0 = Y, W, X_0$
2. **for** $l \in \{1, \ldots, L-1\}$ **do**:
3.    resample $Q$ subsets from $Z_{l-1}$ randomly
4.    **for** $q \in \{1, \ldots, Q\}$ **do**:
5.      **build** tree $T_q$ with subset $q$
6.      **extract** rule set $R_q$ from $T_q$
7.    **generate** leaf encoding $X_l$ with tree from $X_{l-1}$
8.    **combine** $\{Y, W, X_l\} \rightarrow Z_l$
9.    **aggregate** $\{R_1, \ldots, R_q\} \rightarrow R_l$
10. **Output**: $\{Z_1, \ldots, Z_L\}$ and $\{R_1, \ldots, R_L\}$

Fig. 2. Algorithm of Building a Causal Rule Forest

*A. Problem Setup*

In observational studies, not all potential outcomes are observable; we can only observe one outcome per sample while the others remain unobserved. Given the absence of grounded counterfactual outcomes, we adopt Rubin's potential outcomes framework to estimate treatment effects with observational data. The CRF must satisfy the assumption of **unconfoundedness (1)** to ensure unbiased estimations, where the treatment assignment $W$ is independent of the potential outcomes given the covariates $X$:

$$W \perp \{Y(1), Y(0)\} \mid X \quad (1)$$

This critical assumption addresses the challenges of controlling for confounders in a non-randomized setting, positing that all confounders are observable and measurable within the covariates, even if they may be latent variables.

The CRF also relies on the following assumptions to approximate the conditions of randomized trials:

- **Stable Unit Treatment Value Assumption (SUTVA)**

  $Y(t) = Y(t_i)$ for all $t \in \{0, 1\}$ and for all individuals $i$   (2)

  The treatment effect for one individual is independent of the treatment status of other individuals, and there is only one version of the treatment. This encompasses two components, no interference and consistency, ensuring that the treatment effect is well-defined for each individual.

- **Overlap (or Common Support)**

  $0 < P(W=1 \mid X) < 1$ for all $X$   (3)

  For every set of covariates $X$, there is a positive probability of receiving both the treatment and control.

- **Strong Ignorability**

  A combination of unconfoundedness and overlap, posits that, given the same characteristics $X$, the assignment of treatment $W$ and the potential outcomes are independent, and the treatment assignment is not deterministic for given characteristics $X$.

*B. Building Causal Rule Forest*

The CRF fundamentally builds on the algorithm of Causal Trees, emphasizing the principle of honest estimation. This principle is implemented using two distinct data subsets: one for developing the tree structure and another for performing estimation calculations. Such a dual-subset approach is key in identifying potential confounders among observed covariates, enabling an unbiased estimation of the CATE. By training Causal Trees with various subsets created through resampling, the CRF can explore a wider range of treatment heterogeneity. This strategy allows the model to capture a broader spectrum of treatment heterogeneity and enables the detection of more subtle confounding relationships in shallower nodes, which are supported by larger sample sizes, rather than solely in the deeper nodes of causal trees trained on the original datasets.

The training process of CRF unfolds as follows: Consider an N-sample data frame consisting of treatment $W$, $M$ covariates ($X_0$ as referred to Algorithm 1 in Figure 2), and outcome $Y_f$. As depicted in Figure 1, the initial step involves training the first layer of the CRF using this original data frame. In this layer, each tree within the forest is trained on a uniquely resampled subset of the data. Following the training of these trees, we generate leaf encodings for all $Q$ trees within the forest. These leaf encodings, representing new covariates, serve as intermediate rule-encoded representations. Upon completing the training of one forest layer, the CRF outputs a new data frame. This frame encompasses the treatment $W$, outcomes $Y_f$, and a new covariate matrix that includes the leaf encodings from the forest's trees. This newly formed data frame is then utilized to train the subsequent forest layer within the CRF, as outlined in Algorithm 1 in Figure 2. This iterative process of training and generating new data frames with leaf encodings continues layer by layer within the CRF architecture.

IV. EXPERIMENTS

In this section, we first compare the performance of our CRF with other tree-based causal effect estimators on two benchmark datasets: the Infant Health and Development Program (IHDP) [23] and the Jobs dataset [24]. Focusing on supporting decision-making, we include direct estimators such as Causal Trees (CT), Causal Forests (CF), and Causal Trees combined with CRF (CRF + CT), along with an indirect estimator, Bayesian Additive Regression Trees (BART) [25]. We then compare the rules discovered by CT and CRF + CT using data from the Taiwan Longitudinal Study on Aging (TLSA) [26]. The IHDP dataset, introduced by Hill [9], is used to estimate the heterogeneous treatment effects of specialist home visits on cognitive test scores with ML models. By generating potential outcomes using the "A" setting of the NPCI package [27], we use the simulated outcomes as the ground truth of the treatment effects. The Jobs dataset combined observational data with data from a randomized trial [28], and aims to explore the treatment effect of job training on employment status and income. We utilize 100 repetitions of the IHDP and 10 repetitions of the Jobs datasets generated by [6], available at https://www.fredjo.com/.

TABLE I. RESULTS ON IHDP AND JOBS DATASET. LOWER IS BETTER

| Model | Train | | | | Test | | | |
|---|---|---|---|---|---|---|---|---|
| | IHDP | | Jobs | | IHDP | | Jobs | |
| | PEHE | $\epsilon ATE$ | $R_{pol}$ | $\epsilon ATT$ | PEHE | $\epsilon ATE$ | $R_{pol}$ | $\epsilon ATT$ |
| BART | 1.32 ± .36 | .16 ± .04 | .29 ± .03 | .02 ± .01 | **1.40 ± .52** | **.21 ± .05** | .31 ± .03 | .08 ± .06 |
| CT | 3.35 ± .87 | .41 ± .05 | .23 ± .02 | .02 ± .01 | 4.03 ± 1.26 | .62 ± .16 | .25 ± .03 | .09 ± .05 |
| CF[a] | 3.03 ± .88 | .15 ± .03 | .26 ± .02 | .02 ± .01 | 3.40 ± 1.11 | .43 ± .17 | .28 ± .03 | .07 ± .05 |
| CRF + CT[b] | 3.31 ± .88 | .32 ± .04 | .22 ± .01 | .01 ± .01 | 3.83 ± 1.15 | .52 ± .16 | **.24 ± .04** | **.07 ± .04** |

[a.] split.Bucket=FALSE, number of covariates=25, nodesize=1, number of trees=500
[b.] split.Bucket=TRUE, number of covariates=1, nodesize=1, number of trees=200

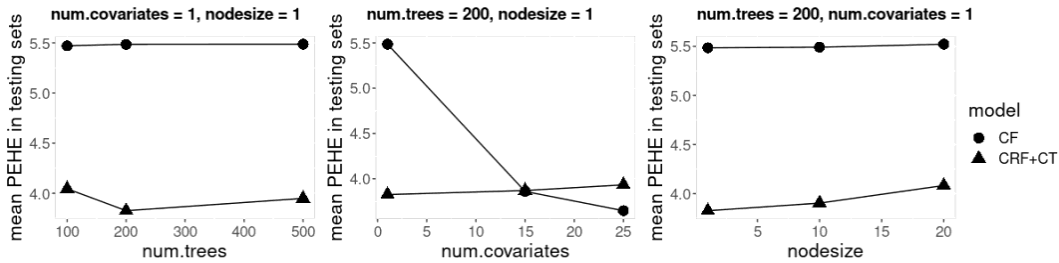

Fig. 3. Performance comparison across different hyperparameters in 100 repetitions of IHDP.

The TLSA dataset, derived from a comprehensive longitudinal survey on aging in Taiwan, commenced in 1989 and has been tracking participants across three cohorts every three or four years. The survey encompasses a broad spectrum of data, ranging from demographic characteristics to emotional support and instrumental abilities. This study specifically examines the effects of self-reported vision problems on scores from the Short Portable Mental Health Questionnaire (SPMSQ), a tool used for detecting potential cognitive impairment [29]. SPMSQ scores, which range from 0 to 10, inversely indicate the severity of cognitive impairment; lower scores signify more severe impairment. Our analysis utilizes data from the 2003 survey (N=4749), a part of the five-wave (1999, 2003, 2007, 2011, 2015) encoded dataset presented by Chien et al. [30]. Within our selected subset, 640 samples (13.4% of the total) with vision problems are identified as the treated group, with mean SPMSQ scores of 9.6 in the control group and 8.4 in the treated group.

### A. Performance on Treatment Effect Estimation

To compare the performance of the proposed method and other estimators, we use four different performance metrics: expected Precision in Estimation of Heterogeneous Effect (PEHE), error on Average Treatment Effect ($\epsilon ATE$), policy risk ($R_{pol}$) [6], and error on Average Treatment Effect on the Treated ($\epsilon ATT$) [31]. We use the 100 repetitions of the IHDP with 90/10 train/test splits and 10 repetitions of Jobs with 80/20 train/test splits, as in previous studies by Shalit et al. [6]. We compare the performance of estimating treatment effects over four estimators with optimal hyper-parameters searched in our experiment. We report the mean value of each measurement with a 95 percent confidence interval in Table 1.

In Table 1, we can see that our CRF + CT outperforms the CT but underperforms the BART and CF. This result shows that training a single CT with representations learned by CRF can improve its performance compared to using raw features in datasets. As the CF is an ensemble model of causal trees, it takes advantage of repeated sampling to learn more diverse information from data. Thus, it is predictable that the performance of CF is better than that of a merely single causal tree. Despite BART having the lowest error in IHDP with a narrower confidence interval, we consider CT as the baseline and BART as an additional comparison. That is because CT-based models and BART use different estimation strategies, and the data-generating processes of synthetic datasets might favor specific algorithms and estimation strategies [13]. In Jobs, our CRF + CT outperforms the other estimators in terms of policy risk and error on average treatment effect on the treated. This shows that our CRF + CT has the potential to outperform the ensemble models and additive models, which are hard to interpret.

To examine the impact of hyper-parameters on the performance of CRF and CT trained with encodings from CRF, we compared CF and CRF + CT across different sets of hyper-parameters based on the optimal hyper-parameters of the CRF + CT presented in Table 1 (split.Bucket=TRUE, number of covariates=1, nodesize=1, number of trees=200). Adjustments were made to the number of trees (num.trees), in-sample covariates (num.covariates), and minimum sample size in leaf nodes (nodesize) as depicted in each segment of Figure 3, aligning CF hyperparameters with those of CRF + CT to serve as a baseline for CRF's capacity.

Our analysis reveals three key insights for parameter tuning when constructing CRF models:

- **Stability Across Tree Quantities**: The results suggest that after a certain threshold, adding more trees does not significantly enhance the precision of treatment effect estimates. This insight is crucial for computational efficiency, indicating that a CRF model with fewer trees could be as effective as its more complex counterpart.

- **Sensitivity to covariates**: We observe that CRF + CT's performance marginally improves with an increase in the number of covariates, unlike CF, which shows a more noticeable improvement. This suggests that the performance of CTs trained with encodings from CRF is less influenced by the number of covariates, as CTs are adept at learning from the interactions among input features. Practically, this implies that using a larger number of covariates in training CRF may not necessarily exacerbate the overfitting issue in models trained with CRF encodings.

- **Node Size and Model Complexity**: The third graph indicates that CTs trained with CRF encodings are more responsive to the size of leaf nodes within the trees of CRF, as opposed to CRF itself. This offers an insight that training CRF under fewer constraints on model complexity may facilitate the learning of more nuanced patterns, thereby enhancing the performance of CTs.

*B. Subgroup Analysis with Causal Rule Forest*

To explore how the rule-encoded representations learned by the CRF can enhance the performance of causal trees with causal data representations. Initially, we constructed both a CT and a CRF + CT with TLSA data and the same setting in tree depth, determined by the minimum number of samples in leaf nodes (minsize=40). We also applied post-pruning based on the minimum cross-validation error of the trees. The leaf node with the largest effect size identified by the CT on the raw TLSA data produces the following rule:

IF bronchitis = 0 ∧ working = 0 ∧ sleeping problems < 1 ∧
family members disputed what they did < 3

THEN CATE = -2.47

This rule suggests that for a group of elders who do not have bronchitis, are not working, do not have sleeping problems, and have occasional family disputes, the SPMSQ score may decrease by 2.47 if they have vision problems. However, such rules, relying solely on simple logical conjunctions (AND), possess limited expressive power compared to rules in multi-level logic.

For instance, the rule from the leaf node with the largest effect size (in terms of absolute treatment effects) in CRF + CT is more complex. Since the CT is trained with rule encodings from CRF, it partitions data into subgroups using the indices of trees as features while indices of rule-encoded regions in each tree are values of the feature:

IF (tree_93 = R1) ∧ ( (tree_32 = R1) ∨ (tree_32 = R6) )

THEN CATE = -3.08

The rule in CNF (AND-of-ORs) involves selected branches and corresponding rules from the forest. When expanded to include original variables for better understanding, the rule becomes:

IF (bronchitis = 0 ∧ ¬ elementary ∧ eye diseases = 0 ∧
Religious activities ≥ 2.5) ∧
( (bronchitis = 0 ∧ heart diseases = 0 &
cannot sleep well < 0.5 ∧ ¬ elementary ∧
Religious activities ≥ 2.5) ∨
(bronchitis= 0 ∧ heart disease =1) )

THEN   CATE = -3.08

Further, these CNF rules can be converted into logically equivalent DNF (OR-of-ANDs), allowing logic optimization to reduce the number of terms and literals:

IF (bronchitis = 0 ∧ ¬ elementary ∧ eye diseases = 0 ∧
Religious activities ≥ 2.5 ∧ heart diseases = 0 ∧
cannot sleep well < 0.5) ∨
(bronchitis = 0 ∧ ¬ elementary ∧ eye diseases = 0 ∧
Religious activities ≥ 2.5 ∧ heart diseases =1)

THEN CATE = -3.08

We can see that CRF + CT can better help subgroup discovery with more specific group definitions and provide more personalized causal-effect inference.

In the preceding example, we identified that the heterogeneity of treatment effects can be more precisely captured using intricate rules derived from the CRF + CT. However, as the causal trees within the CRF deepen, they tend to produce more complex rules in CNF, comprising numerous clauses. This complexity poses a challenge, as transforming these rules from CNF to DNF can lead to an exponential increase in the number of terms. To address this issue, we suggest two potential solutions: firstly, the application of logic optimization to streamline the rules learned by the causal trees with CRF, and secondly, the introduction of a ranking system for the trees in the forest, focusing on those that have demonstrated superior performance in capturing meaningful representations.

Overall, CRF can improve the performance of simple but interpretable models, like causal trees, while keeping their interpretability. Since the CRF can learn diverse and informative confounding relationships that vanilla causal trees are hard to do, employing CRF as a causal representation learner is an effective way to enhance model performance. Besides, the representation learned by CRF can be expressed with human-understandable covariates, which guarantees the interpretability of models. When cooperating CRF with CT, the rules extracted from the CT can identify the heterogeneity coupled with the homogeneity of treatment effects, providing a more comprehensive understanding of the treatment effects.

## V. CONCLUSION

In this study, we demonstrated the efficacy of our proposed CRF in the nuanced domain of causal machine learning, specifically in the estimation of HTE. CRF represents an advancement by harmoniously blending interpretability with analytical precision, a feat that addresses longstanding challenges in this field. Through its innovative use of multi-level Boolean rules and rule-encoded data representations, CRF not only deepens the understanding of subgroup dynamics but also refines the precision in treatment effect estimations, as evidenced in our analyses with the IHDP, Jobs, and TLSA datasets.

Looking forward, the potential applications of CRF in sectors like personalized medicine and policy-making are vast and compelling. Its ability to provide nuanced insights and precise decision support in high-stakes scenarios signifies a significant advancement in causal analysis. Future research may explore ways to enhance CRF's scalability and efficiency, broadening its impact across various complex causal inference challenges. The evolution of CRF and related methodologies is poised to substantially enrich our understanding of causal mechanisms, guiding more informed interventions and policies.